\documentclass{article}
\usepackage{spconf,amsmath,graphicx}
\usepackage{breqn}
\usepackage{csquotes}

\graphicspath{
{./figures/}
}

\title{Semantically Invariant Text-to-Image Generation}
\makeatletter
\def\@name{ \emph{Shagan Sah \enskip Dheeraj Peri \enskip Ameya Shringi \enskip Chi Zhang \enskip Miguel Dominguez}  \\ \emph{Andreas Savakis \enskip Ray Ptucha}}
\makeatother

\address{Rochester Institute of Technology\\
        Rochester NY, USA}
\begin{document}
%
\maketitle
%
\begin{abstract}
Image captioning has demonstrated models that are capable of generating plausible text given input images or videos. Further, recent work in image generation has shown significant improvements in image quality when text is used as a prior. Our work ties these concepts together by creating an architecture that can enable bidirectional generation of images and text. We call this network Multi-Modal Vector Representation (MMVR).
Along with MMVR, we propose two improvements to the text conditioned image generation. Firstly, a n-gram metric based cost function is introduced that generalizes the caption with respect to the image. Secondly, multiple semantically similar sentences are shown to help in generating better images.
Qualitative and quantitative evaluations demonstrate that MMVR improves upon existing text conditioned image generation results by over 20\%, while integrating visual and text modalities.
\end{abstract}

\section{Introduction}

Recent success in image captioning \cite{donahue2015long,chen2015learning, karpathy2015deep, xu2015show} has shown that deep networks are capable of providing apt textual descriptions of visual data. In parallel, advances in conditioned image generation \cite{nguyen2017plug,reed2016generative,nguyen2015deep,nguyen2016synthesizing} provide diverse images from a text based prior.
An ambitious goal for machine learning in the vision and language domain is to be able to represent different modalities of data that have the same meaning with a common latent representation.
For example, words like \enquote{baseball} and \enquote{batter}, a sentence describing a baseball game, or image representations of a baseball game all refer to similar concepts. Concepts that are similar would lie close together in this space while dissimilar concepts would lie far apart.  A sufficiently powerful model should be able to store similar concepts in a similar representation or produce any of these realizations from the same latent space. Successfully mapping visual and textual modalities in and out of this latent space would significantly impact the broad task of information retrieval.

In this paper, we propose a cross-domain model capable of converting between text and image.  By modifying the cost function and introducing multiple sentence conditioning, our model, which we call Multi-Modal Vector Representation (MMVR) improves state of the art \cite{nguyen2017plug} by 23.7\% (from 6.71 to 8.30 inception score). 

The contributions of the paper are as follows: 1) We formulate a latent representation based model that merges inputs across multiple modalities; 2) We propose an n-gram based cost function that generalizes better to a text prior; 3) We show improvements in image quality while using multiple semantically similar sentences for conditioning image generation on generalized text; 4) 
To advance qualitative measurement of text-to-visual models, we introduce an object detector based metric, and conduct human evaluations which compare our metric to the standard inception score \cite{salimans2016improved}. 


\section{Related Work} \label{sec:literature_review}

The notion of a latent space where similar points are close to each other is a key principle of metric learning. The representations obtained from this formulation generalize well when the test data has unseen labels. Models based on metric learning have been used extensively in the domain of face verification \cite{schroff2015facenet}, image retrieval \cite{gordo2017end}, person-re-identification \cite{hermans2017defense} and zero-shot learning \cite{socher2013zero}.

\noindent \textbf{Multi-Modal Learning using Vector Representation -- }
Srivastava \textit{et al.} \cite{srivastava2012multimodal} used deep Boltzmann machines to generate tags from images or images from tags. Sohn \textit{et al.} \cite{sohn2014improved} introduced a novel information theoretic objective that was shown to improve deep multi-modal learning for language and vision. Joint learning based on image category was shown in \cite{reed2016learning}. They used joint training for zero-shot image recognition and image retrieval. Sohn \textit{et al.} \cite{ntuple} introduced multi-class N-tuple loss and showed superior results on image clustering, image retrieval and face re-identification. Eisenschtat \textit{et al.} \cite{2layernet} introduced a 2-layer bidirectional network 
to map vectors coming from two data sources by optimizing correlation loss. Wang \textit{et al.} \cite{imtextembeddings} learned joint embeddings of images and text by enforcing margin constraints on training objectives. Recently, Wu \textit{et al.} \cite{wu2017starspace} leveraged this concept to associate data from different modalities.

\noindent \textbf{Conditional Image Generation -- }
Generative Adversarial Networks (GANs) \cite{goodfellow2014generative} are a sub-class of generative models based on an adversarial game. Training a GAN involves two models: a generator that maps a random distribution to the data distribution; and a discriminator that estimates the probability of a sample being fake or real. A GAN can produce sharp images but the generated images are not always photo-realistic. To improve upon photo-realistic quality, class category \cite{nguyen2016synthesizing,nvidiagan,wang2017highres}, caption \cite{nguyen2017plug, reed2016generative} or a paragraph \cite{liang2017recurrent} has been used to condition image generation. Reed \textit{et al.} \cite{reed2016generative} encoded text into a vector to condition images, however direct encoding reduces the diversity of generated images. Introducing an additional prior on the latent code, Plug and Play Generative Networks (PPGN) \cite{nguyen2017plug} drew a wide range of image types and introduced an image conditioning framework.



\section{Methodology} \label{sec:methodology}

\subsection{Multi-Modal Vector Representation} \label{sub:mmvr}
\begin{figure}[!ht]
  \centering
  \centerline{\includegraphics[width=.4\textwidth]{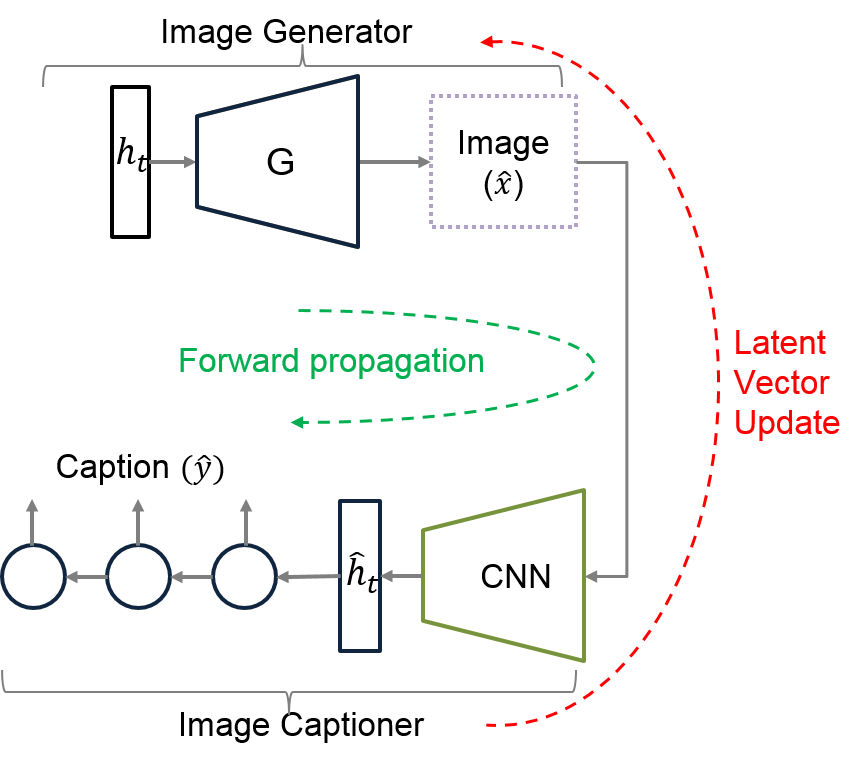}}
  \caption{Overview of the MMVR model. It consists of two pre-trained modules -- an image generator (G) that inputs a latent representation $h_t$ and generates an image $\hat{x}$; and an image captioner that inputs an image $\hat{x}$ and generates a caption $\hat{y}$. To update the latent vector $h_t$, cross-entropy between the generated caption $\hat{y}$ and a ground truth caption $y$ is used while the weights for the generator and CNN are fixed.}
\label{fig:mmvr}
\end{figure}

Inspired by \cite{nguyen2017plug}, we introduce Multi-Modal Vector Representation (MMVR) to create a unified representation for visual and text modality in latent space. 
Figure \ref{fig:mmvr} provides an overview of the MMVR architecture. The model can be divided into two interdependent modules: an image generator based on \cite{dosovitskiy2016generating} and an image captioner based on \cite{donahue2015long}.

The forward pass is initiated by passing a random latent vector $h_t$ into the image generator which generates an image $\hat{x}$. The image captioner uses the generated image to create a caption. Word-level cross entropy is used to determine the error between the generated caption, $\hat{y}$ and a ground truth caption $y$. This error is used to iteratively update $h_t$ (and thus $\hat{x}$), while keeping all other components fixed.  With each iteration, $\hat{y}$ approaches $y$, and the generated image $\hat{x}$ serves as a proxy for the target caption.  The gradient associated with the cross-entropy error is specified in  (\ref{eqn:grad}).

\begin{dmath}
\label{eqn:grad}
grad(C) =\frac{\partial \mathcal{L}(C_{pred}, C_{gt})}{\partial h_{t}}
\end{dmath}

\noindent Where, $grad(C)$ is the gradient of cross-entropy with respect to latent vector $h_{t}$, $C_{pred}$ is the predicted caption and $C_{gt}$ is a ground truth caption. $\mathcal{L}$ is the word level cross-entropy between the two captions.
The $grad(C)$ component of the update rule ensures that the generated images have relevant context. However, to improve the realistic nature of the images, a reconstruction error is included in the update rule. This is computed as the difference between $h_t$ and $\hat{h_t}$. This component is referred to as a Denoising Autoencoder (DAE) in \cite{nguyen2017plug}. Finally, to add diversity in generated images, a noise term $\mathcal{N}$ is also included. The resulting update rule is a weighted sum of four terms and is described in (\ref{eqn:ppgn}). 

\begin{dmath}
\label{eqn:ppgn}
h_{t+1} = h_{t} + \gamma_{1}grad(C) +
\gamma_{2} R(h_{t}, \hat{h_{t}}) + \mathcal{N}(0,\gamma_{3})    
\end{dmath}

\noindent Where, $R(h_{t}, \hat{h_{t}})$ is the reconstruction error which is computed as difference between $h_{t}$ and  $\hat{h_{t}}$, $\mathcal{N}$ is Gaussian noise with standard deviation $\gamma_3$ and $h_{t+1}$ is the latent vector after the update. $\gamma_1$ and $\gamma_2$ are weights associated with the gradient of cross entropy and the DAE, respectively.


\subsection{n-gram Metric Conditioning}
An intrinsic limitation with the model in Section \ref{sub:mmvr} is that the cross-entropy loss requires exact word level correspondences between generated and ground truth captions. For example, consider a case when the generator is conditioned on \enquote{a red car}, whereas the captioner outputs \enquote{the car is red}. Both captions are semantically very similar but lack one-to-one correspondence between words. This may result in unwanted updates of the latent vector $h_t$ due to high word level cross-entropy. We address this by introducing a n-gram metric in the latent vector update. Our metric is responsive to cases when generated and reference captions are different, but semantically similar.

Equation (\ref{eqn:bleu}) describes the updated $\gamma_1$ term when the n-gram metric is used in conjunction with cross-entropy. We compute word level differences and scale $\gamma_1$ with the n-gram metric between the generated and reference captions:



\begin{dmath}
\label{eqn:bleu}
\gamma_{1}\frac{1-\mathcal{F}(C_{pred}, C_{gt})}{n}grad(C)
\end{dmath}

\noindent Where $\mathcal{F}$ is the n-gram metric. In our experiments, we use the BLEU \cite{papineni2002bleu} scores as n-gram metric. As before, our latent vector $h_t$ is obtained through an iterative process.  When the captions are semantically similar, the magnitude of the update is significantly reduced by n-gram metric scaling, preventing unwanted updates to the latent vector.

\subsection{Conditioning on Multiple Captions}
Another way to overcome one-to-one word correspondences between a predicted and reference sentence is to use semantically similar sentences.

Multiple captions would increase syntactic variability for the generator to condition on, hence improving the overall image quality.
The forward pass is performed in a same way as Section \ref{sub:mmvr}. The predicted caption is compared against multiple captions from a sentence paraphraser \cite{zhang2017semantic} to obtain the individual gradients. The aggregated gradients are used to update the latent vector $h_t$. The caption gradient component of the $h_t$ update rule is replaced by the summation of gradients from multiple captions as $grad_{avg} = \frac{1}{N_C} \sum_{i=1}^{N_C} grad(C_i)$, where, $N_{C}$ is the total number of reference captions and $grad_{avg}$ is the aggregated gradient for all captions.

\section{Results and Discussion} \label{sec:experimental_results}


\subsection{Inference}
We start with a random 4096-dimensional vector $h_t$ to render a 256$\times$256 image and iteratively update $h_t$.
The process is terminated after 200 iterations and the resulting image is treated as a representative image for the caption. We set $\gamma_1$ and $\gamma_2$ hyper-parameters from (2) as 1 and $10^{-3}$, respectively.

\subsection{Evaluation Metrics}

\begin{figure}[!ht]
  \centering
  \centerline{\includegraphics[width=.48\textwidth]{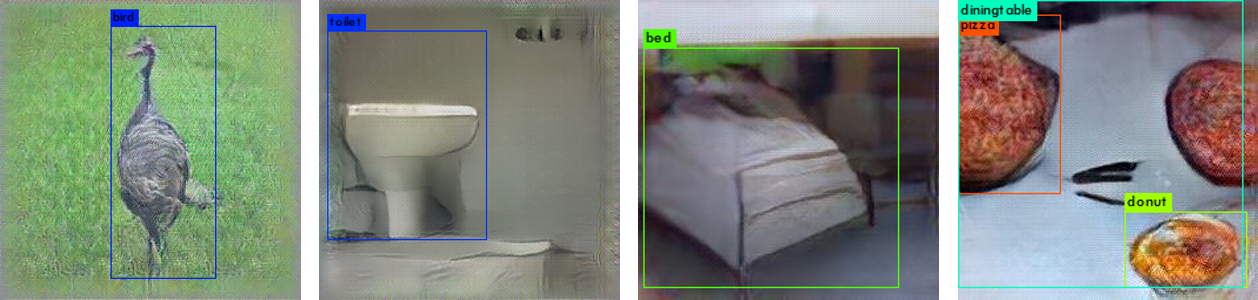}}
  \caption{Examples of the YOLO object detection on generated images. The bounding boxes and corresponding labels are detections with confidence greater than $0.5$ threshold.}
\label{fig:yolo}
\end{figure}

We evaluate image generation tasks through qualitative comparisons as well as by quantitative metrics and human evaluations.
In addition to using the inception score \cite{salimans2016improved} metric, we propose a new metric based on object detection that captures the quality of multiple objects present in a generated image.
A pre-trained YOLO object detector model \cite{redmon2016yolo9000} is used for this purpose. The model is trained on 80 object categories commonly present in the MS-COCO dataset. We show some examples in Figure \ref{fig:yolo} with synthesized images. Each synthesized image is passed through the object detector model that yields bounding boxes and their corresponding confidences. Formally, $ detection~score = \sum_{d} \frac{A_d}{A_T} p_d$, which reports the weighted sum of all detections ($d$) greater than a $0.1$ confidence threshold ($p_d$), where the weight is the ratio of the detected bounding box area ($A_d$) and the full image area ($A_T$). Having an area weight is critical since some object detector models may predict a large number of small bounding boxes. 

\noindent{\textbf{Human Evaluations -- }}
We conduct human evaluations to validate image generation.
We collected 50 image-caption pairs and asked 80 humans (not including any of the authors) to judge the performance. Each participant was shown eight random images from all methods in random order totaling to 40 samples per person. Each evaluator was asked to rate on a 1 (bad) $-$ 5 (good) Likert-type scale. On average, each method received more than 600 ratings. The questions asked to the human judges were: (1) Can you identify any one object in the image? and (2) How well does the sentence align with the image?



\subsection{Text-to-Visual}
\begin{figure}[!ht]
  \centering
  \centerline{\includegraphics[width=.4\textwidth]{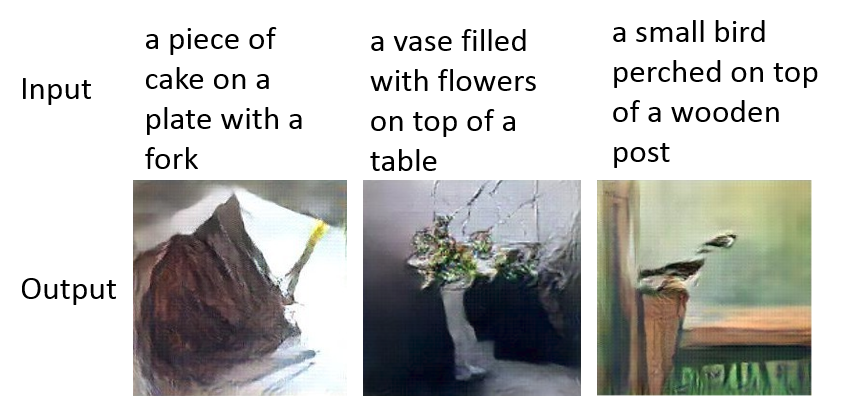}}
  \caption{Examples of text-to-visual transformation.}
\label{fig:img_gen}
\end{figure}

Cross-modal experiments aid in proving that the representations of individual modalities are well aligned in the common space. We show examples of text-to-visual generation in Figure \ref{fig:img_gen}. It can be observed that MMVR synthesizes reasonable images from captions. As noted in \cite{nguyen2017plug}, one of the major challenges while conditioning on text include the cross-entropy computation from a sentence with many words. The captions could be 10-15 words long including stop-words which have limited significance on the image content. Moreover, gradients for all words are aggregated and back-propagated, hence significant words may loose importance. This may result in poor image quality. The inclusion of n-gram scaling to the update function and conditioning on multiple ground truth sentences help address such limitations.

Table \ref{inception_score} compares the text-to-visual techniques against a baseline (direct FC-6). The inception scores indicate the improvement in generated images when  BLEU-1 (B-1) and the multiple caption conditioning ($N_c=5$) are used. The detection scores for multiple captions are significantly better than other variants. However, BLEU-1 is slightly lower than the baseline result. Despite having worse inception scores, baseline methods got higher human evaluation scores. We believe the reason for this trend is lack of detail in objects generated by multiple captions. The baseline model generates images with single objects, making them visually appealing.

\begin{table}[!ht]
\centering
\caption{Evaluation of the generated image quality using the inception, detection and human scores on the test set.}
\label{inception_score}
\begin{tabular}{c|c|c|c}
\hline
Method           & Inception & Detection & Human \\ \hline 
Baseline (FC-6)  & $5.77 \pm 0.96 $  & 0.762 & \textbf{2.95} \\ 
PPGN \cite{nguyen2017plug} & $6.71 \pm 0.45 $  & 0.717 & 2.34 \\  
MMVR (B-1) & $7.22 \pm 0.81 $ & 0.713 & 2.31    \\ 
MMVR ($N_c=5$) & $\textbf{8.30} \pm \textbf{0.78}$  & \textbf{1.004} & 2.71 \\ \hline
\end{tabular}
\end{table}

\noindent{\textbf{Conditional Image Generation on Multiple Sentences -- }
Synthetic sentences were generated using a sentence paraphraser \cite{zhang2017semantic}. Figure \ref{fig:text2img_mult_sent} shows the input caption and the generated images with 1, 3 and 5 captions. Image quality enhances with increase in number of sentences. The \emph{food} example also show gains in understanding the concept of quantity (\emph{four}) through text. Similar trends are observed through the inception and detection score metrics as reported in Table \ref{input_sent_inception_score}. The detection score helps prove that multiple sentences assist in generating multiple objects in the image that are recognized by the object detector. 

\begin{figure}[!ht]
  \centering  \centerline{\includegraphics[width=.42\textwidth]{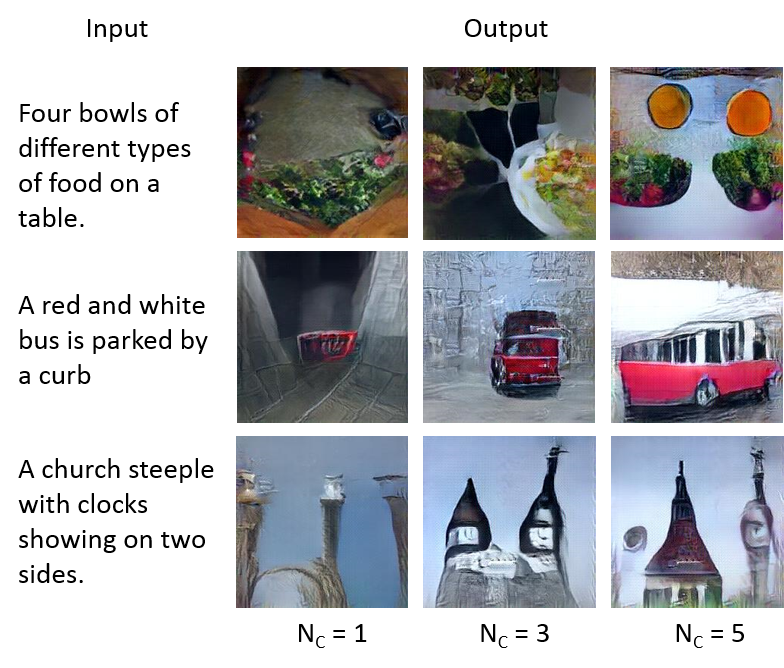}}
  \caption{Examples of the text-to-image generation as conditioned on varying number of input captions. We observe more detailed images being synthesized with increase in captions.}
\label{fig:text2img_mult_sent}
\end{figure}

\begin{table}[!ht]
\centering
\caption{Evaluation of the generated image quality by conditioning on varying number of paraphrased sentences ($N_C$).}
\label{input_sent_inception_score}
\begin{tabular}{c|c|c|c}
\hline
$N_C$ & Inception & Detection & Human \\ \hline 
1   & $7.22 \pm 0.81 $  & 0.713 & 2.30 \\ 
3   & $8.04 \pm 0.57 $  & 0.915 & \textbf{2.73}\\ 
5   & $\textbf{8.30} \pm \textbf{0.78} $  & \textbf{1.005} & 2.71 \\ \hline
\end{tabular}
\end{table}

\noindent{\textbf{Was the n-gram scaling useful ? }}
We show examples with and without the n-gram scaling of the gradient term in (3) in Figure \ref{fig:ppgn_comp}. It is very difficult to judge the two techniques visually. We use only a single caption to condition the image generator to have a fair comparison in this case. The BLEU-1 score was used as the word level error multiplier and it scales the gradients accordingly. The inception scores in Table \ref{inception_score} show slight improvement for BLEU-1 against the PPGN.

\begin{figure}[!ht]
  \centering
  \centerline{\includegraphics[width=.38\textwidth]{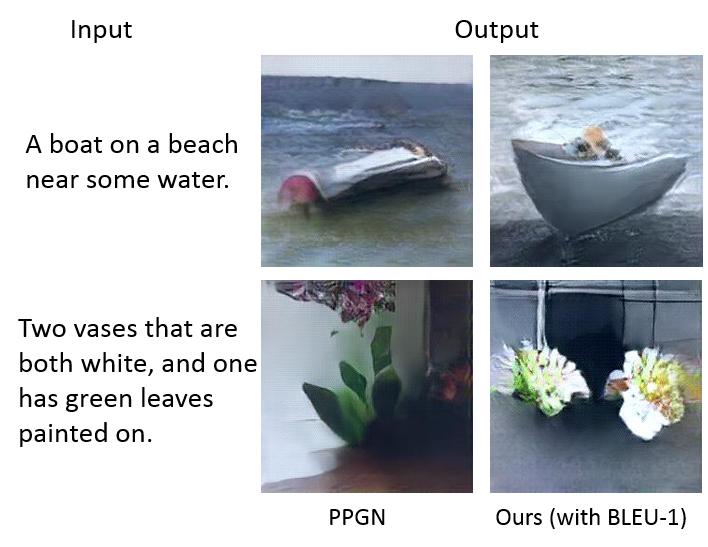}}
  \caption{Examples comparing the text-to-image for PPGN and the BLEU-1 scaled cross-entropy. Even though slight improvements could be observed with the n-gram scaling, judging the image quality visually is very challenging.}
\label{fig:ppgn_comp}
\end{figure}

\noindent{\textbf{Which degree n-gram is better for scaling ? }}
We compare different BLEU scaling in (3) by varying the n-gram metric. Results are reported in Table \ref{bleu_inception_score}. One reason the BLEU-1 performs better than the higher n-gram techniques might be the simple removal of one-to-one word correspondences between the predicted and ground truth captions is sufficient. Higher BLEU metrics require n-gram matching which puts hard constraints on the generated caption. This may dampen the significance on important words in the overall update.

\begin{table}[!ht]
\centering
\caption{Comparison of image quality with different BLEU metrics for scaling the latent vector update function.}
\label{bleu_inception_score}
\begin{tabular}{c|c}
\hline
Scaling n-gram Metric & Inception Score \\ \hline 
BLEU-1 & $ \textbf{7.22} \pm \textbf{0.81} $  \\ 
BLEU-2 & $ 7.12 \pm 0.66 $  \\ 
BLEU-3 & $ 7.05 \pm 0.73 $  \\
BLEU-4 & $ 6.83 \pm 0.74 $  \\ \hline
\end{tabular}
\end{table}


\noindent{\textbf{Does fine-tuning the image generator help ? -- }}
The generator was unable to address common words that occur in a caption (man, woman, person, \emph{numbers}, etc.) since ImageNet does not contain such categories. Moreover, some dominant categories in MS-COCO dataset like giraffe, stop sign and person are not present in ImageNet dataset.
By fine-tuning, the generator is able to semantically capture such categories.
\section{Conclusion}
\label{sec:conclusion}
This work advances the area of caption conditioned image generation by allowing the vector space to be shared between vision and language representations. MMVR shows flexibility in performing cross-modal transformations and improves state-of-the-art by more than 20\%. We address some limitations such as one-to-one word correspondence by using a n-gram metric and conditioning on multiple semantically similar sentences. We introduce a new objective metric for evaluating generated images which allows for multiple objects per generated image. 


\bibliographystyle{IEEEbib}
\bibliography{egbib}

\begin{thebibliography}{10}

\bibitem{donahue2015long}
Jeffrey Donahue et~al.,
\newblock ``Long-term recurrent convolutional networks for visual recognition
  and description,''
\newblock in {\em Proceedings of the IEEE conference on CVPR}, 2015, pp.
  2625--2634.

\bibitem{chen2015learning}
Xinlei Chen and C~Lawrence Zitnick,
\newblock ``Learning a recurrent visual representation for image caption
  generation,''
\newblock 2015.

\bibitem{karpathy2015deep}
Andrej Karpathy and Fei-Fei Li,
\newblock ``Deep visual-semantic alignments for generating image
  descriptions,''
\newblock in {\em Proceedings of the IEEE CVPR}, 2015, pp. 3128--3137.

\bibitem{xu2015show}
Kelvin Xu et~al.,
\newblock ``Show, attend and tell: Neural image caption generation with visual
  attention,''
\newblock {\em arXiv preprint arXiv:1502.03044}, vol. 2, no. 3, pp. 5, 2015.

\bibitem{nguyen2017plug}
Anh Nguyen et~al.,
\newblock ``Plug \& play generative networks: Conditional iterative generation
  of images in latent space,''
\newblock in {\em CVPR, 2017 IEEE Conference on}. IEEE, 2017, pp. 3510--3520.

\bibitem{reed2016generative}
Scott Reed et~al.,
\newblock ``Generative adversarial text to image synthesis,''
\newblock {\em arXiv preprint arXiv:1605.05396}, 2016.

\bibitem{nguyen2015deep}
Anh Nguyen, Jason Yosinski, and Jeff Clune,
\newblock ``Deep neural networks are easily fooled: High confidence predictions
  for unrecognizable images,''
\newblock in {\em Proceedings of the IEEE Conference on CVPR}, 2015, pp.
  427--436.

\bibitem{nguyen2016synthesizing}
Anh Nguyen et~al.,
\newblock ``Synthesizing the preferred inputs for neurons in neural networks
  via deep generator networks,''
\newblock in {\em NIPS}, 2016, pp. 3387--3395.

\bibitem{salimans2016improved}
Tim Salimans et~al.,
\newblock ``Improved techniques for training gans,''
\newblock in {\em Advances in Neural Information Processing Systems}, 2016, pp.
  2234--2242.

\bibitem{schroff2015facenet}
Florian Schroff, Dmitry Kalenichenko, and James Philbin,
\newblock ``Facenet: A unified embedding for face recognition and clustering,''
\newblock in {\em Proceedings of the IEEE Conference on CVPR}, 2015, pp.
  815--823.

\bibitem{gordo2017end}
Albert Gordo et~al.,
\newblock ``End-to-end learning of deep visual representations for image
  retrieval,''
\newblock {\em International Journal of Computer Vision}, vol. 124, pp.
  237--254, 2017.

\bibitem{hermans2017defense}
Alexander Hermans, Lucas Beyer, and Bastian Leibe,
\newblock ``In defense of the triplet loss for person re-identification,''
\newblock {\em arXiv preprint arXiv:1703.07737}, 2017.

\bibitem{socher2013zero}
Richard Socher et~al.,
\newblock ``Zero-shot learning through cross-modal transfer,''
\newblock in {\em Advances in neural information processing systems}, 2013, pp.
  935--943.

\bibitem{srivastava2012multimodal}
Nitish Srivastava and Ruslan~R Salakhutdinov,
\newblock ``Multimodal learning with deep boltzmann machines,''
\newblock in {\em Advances in neural information processing systems}, 2012, pp.
  2222--2230.

\bibitem{sohn2014improved}
Kihyuk Sohn, Wenling Shang, and Honglak Lee,
\newblock ``Improved multimodal deep learning with variation of information,''
\newblock in {\em Advances in Neural Information Processing Systems}, 2014, pp.
  2141--2149.

\bibitem{reed2016learning}
Scott Reed et~al.,
\newblock ``Learning deep representations of fine-grained visual
  descriptions,''
\newblock in {\em Proceedings of the IEEE Conference on CVPR}, 2016, pp.
  49--58.

\bibitem{ntuple}
Kihyuk Sohn,
\newblock ``"improved deep metric learning with multi-class n-pair loss
  objective,''
\newblock in {\em NIPS}, 2016.

\bibitem{2layernet}
Aviv Eisenschtat and Lior Wolf,
\newblock ``Linking image and text with 2-way nets,''
\newblock in {\em arXiv preprint arXiv:1608.07973}, 2016.

\bibitem{imtextembeddings}
Svetlana Wang, Yin~Li,
\newblock ``Learning deep structure-preserving image-text embeddings,''
\newblock in {\em Proceedings of the IEEE Conference on CVPR}, 2016.

\bibitem{wu2017starspace}
Ledell Wu et~al.,
\newblock ``Starspace: Embed all the things!,''
\newblock {\em arXiv preprint arXiv:1709.03856}, 2017.

\bibitem{goodfellow2014generative}
Ian Goodfellow et~al.,
\newblock ``Generative adversarial nets,''
\newblock in {\em NIPS}, 2014, pp. 2672--2680.

\bibitem{nvidiagan}
T.~Karras, T.~Aila, S.~Laine, and J.~Lehtinen,
\newblock ``Progressive growing of gans for improved quality, stability, and
  variation.,''
\newblock {\em arXiv preprint arXiv:1710.10196}, 2017.

\bibitem{wang2017highres}
Ting-Chun Wang et~al.,
\newblock ``High-resolution image synthesis and semantic manipulation with
  conditional gans,''
\newblock {\em arXiv preprint arXiv:1711.11585}, 2017.

\bibitem{liang2017recurrent}
Xiaodan Liang et~al.,
\newblock ``Recurrent topic-transition gan for visual paragraph generation,''
\newblock {\em arXiv preprint arXiv:1703.07022}, 2017.

\bibitem{dosovitskiy2016generating}
Alexey Dosovitskiy and Thomas Brox,
\newblock ``Generating images with perceptual similarity metrics based on deep
  networks,''
\newblock in {\em NIPS}, 2016, pp. 658--666.

\bibitem{papineni2002bleu}
Kishore Papineni et~al.,
\newblock ``Bleu: a method for automatic evaluation of machine translation,''
\newblock in {\em Proceedings of the 40th annual meeting on association for
  computational linguistics}. Association for Computational Linguistics, 2002,
  pp. 311--318.

\bibitem{zhang2017semantic}
Chi Zhang et~al.,
\newblock ``Semantic sentence embeddings for paraphrasing and text
  summarization,''
\newblock in {\em GlobalSIP}, 2017.

\bibitem{redmon2016yolo9000}
Joseph Redmon and Ali Farhadi,
\newblock ``Yolo9000: better, faster, stronger,''
\newblock {\em arXiv preprint arXiv:1612.08242}, 2016.

\end{thebibliography}

\end{document}